\documentclass{llncs}
\usepackage{llncsdoc}
\usepackage{algorithm}
\usepackage{algorithmic}
\usepackage{setspace}
\usepackage{graphicx}
\usepackage{cite}
\usepackage{booktabs}
\usepackage{caption}
\usepackage{subfigure}
\usepackage{epstopdf}
\usepackage{epsfig}
\usepackage{pdfpages}

\begin{document}
\title{Integrating Feature and Image Pyramid: \\A Lung Nodule Detector Learned in Curriculum Fashion}
\author{Benyuan Sun\inst{1, 2} \and Zhen Zhou\inst{2} \and Fandong Zhang\inst{2} \and Xiuli Li\inst{2} \and Yizhou Wang\inst{1, 2}}
\institute{Nat'l Engineering Laboratory for Video Technology,\\ Key Laboratory of Machine Perception (MoE),\\
    Cooperative Medianet Innovation Center, Shanghai\\
	 Sch'l of EECS, Peking University, Beijing, China
\and
Deepwise Inc., Beijing, China}
\maketitle
\begin{abstract}
Lung nodules suffer large variation in size and appearance in CT images. Nodules less than 10mm can easily lose information after down-sampling in convolutional neural networks, which results in low sensitivity. In this paper, a combination of 3D image and feature pyramid is exploited to integrate lower-level texture features with high-level semantic features, thus leading to a higher recall. However, 3D operations are time and memory consuming, which aggravates the situation with the explosive growth of medical images. To tackle this problem, we propose a general curriculum training strategy to speed up training. An dynamic sampling method is designed to pick up partial samples which give the best contribution to network training, thus leading to much less time consuming. In experiments, we demonstrate that the proposed network outperforms previous state-of-the-art methods. Meanwhile, our sampling strategy halves the training time of the proposal network on LUNA16.
\end{abstract}

\section{Introduction}
Lung cancer is one of the leading causes of cancer death worldwide\cite{national2011reduced}. Recent researches show Computed Tomography (CT) can help diagnose lung nodule in an early stage. To relieve the heavy burden of radiologists, computer-aided detection systems have been developed. Recent competitions like LUNA16\cite{DBLP:journals/corr/SetioTBBBC0DFGG16} and Kaggle Data Science Bowl 2017 further promote development in this area. However, there still remains some challenges on this task:
\begin{alpherate}
\item
Lung nodules suffer from large variations in shape, size and appearance. This produces totally different images for the detector thus the model's robustness to these variation is required.
\item
Large inter \& intra class imbalance exist in medical data. Limited nodule data of particular type makes it hard for networks to capture the discriminative features.
\item
With the increasing growth of medical data, efficient training schedule is required. For example, a typical detection network based on 3D convolutions needs about one day to train on LUNA16 dataset. Time consuming will become a big problem when the data size further increases.
\end{alpherate}


Deep Convolutional Neural Network (DCNN) is explored to extract discriminative features. Ding et al.\cite{ding2017accurate} first proposed to use 2D convolutional network to generate nodule proposals and a 3D ConvNet for nodule classification. Dou et al.\cite{dou2017automated} use two stage 3D ConvNets and Hybrid-Loss to overcome the data imbalance. However, these works failed to capture the large variance of nodule size, which is of great importance in nodule detection.

In this paper, we develop a novel two-stage nodule detector that integrates both image and feature pyramid for nodule detection. Firstly, to avoid the detail information missing in upper layers of DCNN, we extend Feature Pyramid Network (FPN)\cite{lin2017feature} to 3D as our nodule proposal network. Given one CT as input, FPN generates rich semantics feature maps at different resolution by fusing both high and low features, enabling nodule detection in the proper resolution. Secondly, a image pyramid is designed for further false positive reduction. Due to lack of knowledge of object's size, traditional image pyramid consists of a set of images with different scales\cite{adelson1984pyramid}. In this paper, we use FPN to produce a rough size information of the proposal and detect the proposals in a proper resolution.

However, 3D FPN brings heavy computation burden and greatly increases the training time. To solve this problem, we develop an intuitive curriculum learning strategy that significantly boosts the training procedure. Curriculum learning is an idea proposed by Bengio et al.\cite{bengio2009curriculum} that learning organized in a 'meaningful order' both speeds up and outperforms than those data are randomly sampled. Graves et al.\cite{graves2017automated} models the training procedure as a non-stationary multi-armed bandit and use exp3 algorithm to solve this problem. Jesson et al.\cite{10.1007/978-3-319-66179-7_73} proposed an curriculum that gradually changes from nodule surroundings to the whole image. Inspired by \cite{graves2017automated}, data with high loss (high gradient) or less-trained are regarded as 'meaningful', which can boost training. Moreover, in this paper, we provide both theoretical and experimental analysis in terms of the efficiency of the proposed method.

\section{Our Proposed Detection Model}
Our nodule detection system is a typical two-stage detector. The first stage takes the whole CT as input and outputs nodule proposals, achieving high sensitivity but remaining high false positive rates. The second module reduce false positives generated by the previous stage to increase specificity. We integrates FPN and Image Pyramid in an computation-efficient manner.
\subsection{Nodule Proposal Network}
Nodule Proposal Network takes the whole CT as input and outputs a set of nodule proposals with confidence score. Each proposal consists a coordinate $(x, y, z)$ in 3d space representing its position in CT and a diameter $r$ stands for its size. This process is modeled by a improved 3d Feature Pyramid Network (FPN).

\begin{figure}[ht]
\centering
\includegraphics[scale=0.2]{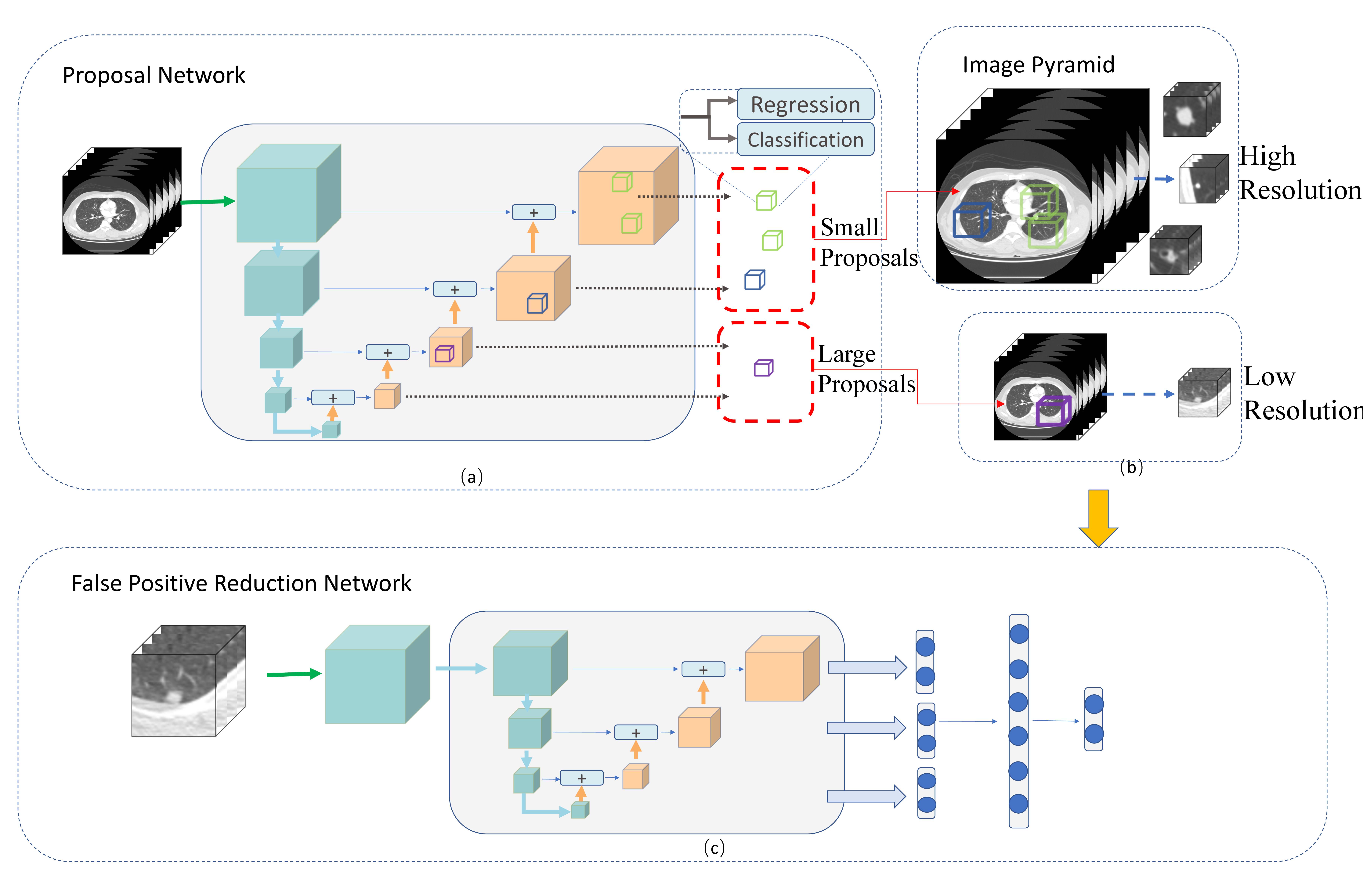}
\caption{The Structure of our proposed Detector}
\label{fig:fpn}
\enspace
\end{figure}

To detect objects in a proper resolution, FPN generates a series of feature map, called feature pyramid (Fig.\ref{fig:fpn} (a) orange parts). Then it assumes a location mapping between the feature maps and original image(Fig.\ref{fig:fpn} (a) color mapping). The probability of one coordinate to be a nodule is given by the classification result at the corresponding anchor (voxel) in the feature map. We convert the original FPN to 3D and add a few more upsampling layers to fit small nodules, see Fig.\ref{fig:fpn}. The biggest feature map in the feature pyramid is the same size as the input image, this enables a strong per-voxel detection.\\

Prediction is done as in \cite{lin2017feature}. We predicts a confidence $s_i$ at every anchor point (every voxel) of each feature map, and regress an offset $t_i$ for each anchor. $t_i$ is a vector representing the 4 parameterized coordinates osf the predicted nodule circle. We use Focal Loss\cite{lin2017focal} for classification and Huber Loss for Regression. Focal loss is the best choice when facing large imbalance in data.

\subsection{False Positive Reduction}
We crop $48\times48\times48$ cubes centered at each proposal from the original CT, and use these cubes as input for a refined classification. But large variation between different size of nodules impacts the training procedure. Here we develop two solutions for this problem: 1. a image pyramid schedule to generate inputs. 2. Feature Pyramid Pooling to capture information in different resolution.

\textbf{Image Pyramid Inputs} Traditional image pyramid repeatedly compute of the same nodule. However, the nodule candidates given by a FPN network posses a rough prior of size: nodule size grows with the depth of feature map. Given the size prior, a proper resolution is assigned to proposal as in Fig. \ref{fig:fpn} (b).

\textbf{Feature Pyramid Pooling} We use a similar network to our proposal network for false positive reduction as in proposal stage. The biggest feature map is discarded to accelerate computing. The feature pooled at different feature map is concated and used as input to a fully connected layer. See Fig. \ref{fig:fpn} (c).

\section{Curriculum Training Strategy}
In this section, we introduce our curriculum learning strategy for proposal network. We assume labels $y_i$ are conditionally modeled on their inputs $x_i$ in our dataset $\mathcal{X} := \{(x_i, y_i)| i\in{1, 2, ..., n}\}$. Now a task $\mathcal{T}$ is a distribution $\mathcal{D}$ over the dataset $\mathcal{X}$. Curriculum learning is to choose a sequence of tasks $\mathcal{T}_1, \mathcal{T}_1, ..., \mathcal{T}_N$ for the classifier(e.g. Neural Networks) to train on, where the goal is to improve either the learning speed or performance on the final task $\mathcal{T}_N$. In this paper, we propose two methods to generate distributions for training data sampling, which is applied on mini-batch of CT and anchors of negative samples respectively.

\subsection{Strategy of Sampling mini-batch}
Most optimization methods for neural networks are based on Loss Function $\mathcal{L}$ and back-propagation, which may waste a lot of time on low loss samples. Sample $(x_i, y_i)$ with low loss $\mathcal{L}(x_i, y_i)$ has gradient close to zero and contributes little to the training procedure. One method to solve this problem is to use hard-negative mining, but this may cause forgetting: model might forget the already learned samples after training on hard samples, and the forgotten data cannot be retrieved again because the loss will not be updated without a re-visiting.

We propose a much smoother sampling inspired by UCB1\cite{auer2002finite}. For each training sample $(x_i, y_i)\in\mathcal{X}$, we maintain a state $(L_i, N_i)$, where $L_i = \sum_{k=N_i-c}^{N_i} \mathcal{L}(x_i, y_i)$ records the average loss of recent $c$ training steps on this sample, and $N_i$ is the training iterations on this sample. Here, task $\mathcal{T}_t$ can be considered as $t$th training epoch, and the weight of a specific sample is calculated by:
\begin{equation}
    w_i := L_i + \frac{\alpha}{\sqrt{N_i}} \label{equ:weight}
\end{equation}
then its distribution function $D_t$ is $\pi_t^i := (1 - \epsilon)e^{w_i}/\sum_{j=1}^n{e^{w_j}} + \frac{\epsilon}{n}\enspace$
, where $\alpha$ is a balance factor between the two losses and $\epsilon$ is a hyper parameter encouraging exploration.

For task $\mathcal{T}_t$, we sample fixed number $N_{epoch}$ of data from the distribution $D_t$ as the training samples. The training procedure as a whole can be viewed in Algorithm \ref{alg:1}. Intuitively, our method can be considered as a balance between hard negative mining and low frequency data training. Forgetting is largely avoided because the forgotten samples will have a relatively larger $\frac{\alpha}{\sqrt{N_i}}$ after several epochs.

\vspace{-0.4cm}
\subsubsection{Theoretical Explanation of our method}
We further give some theoretical explanation to our method. A training step on sample $x_i$ at time $t$ will give a reward $r_{i, t}$ to the network. We suppose that the total reward $R_t = \sum{r_{i_t, t}}$ received by the network has a close relation with it's performance, where $i_t$ is the data chosen at time $t$.

Now our training procedure can be viewed as a Non-stationary Multi-armed Bandit\cite{besbes2014stochastic} where each bandit is one training data. However, the best performance algorithm must incur a regret of at least order $T^{2/3}$ when the reward is non-stationary\cite{besbes2014stochastic}. So we go a few more steps. Noting that although the reward $r_{i, t}$ changes with time, but they all declined to 0 when the model is perfectly trained. We assume the expected reward decline ratio $d_t = E(r_{i, t} / r_{i, t-1})$ after training step $t$ is only related to the model.

\begin{lemma}If $r_{i, t}$ are independent sequences that converges to 0 and expected decline ratio $d_t = E(r_{i, t} / r_{i, t-1})$ is independent with any specific reward at each time step t, then $\hat{r_{i, t}} = r_{i, t}\prod_{j=1}^{t-1}{d_t}$ are variables with same distribution $\mathcal{P}_i$.\end{lemma}

By Lemma. 1, we can easily get a new reward $\hat{r_{i, t}}$ that has the same distribution $P_i$ by re-scaling the rewards at each time step. The new reward formulates a stationary multi-armed bandit problem. By the proof of UCB1 algorithm\cite{auer2002finite}:

\begin{theorem}For all $k > 1$, if policy UCB1(always choosing largest $w_i$) is run on K machines having arbitrary reward distributions $P_1,..., P_k$ with support in $[0, 1]$, then its expected regret $R$ after $n$ plays is at most:
\begin{equation}
    R = 8 \times \sum_{i:\mu_i<\mu^*}(\frac{\ln n}{\delta_i}) + (1 + \frac{\pi^2}{3}) \times \sum_{j=1}^k{\delta_j}
\end{equation}
where $\mu_1, ..., \mu_k$ are expected values of $P_1, ..., P_k$ and $\delta_i = \mu^* - \mu_i$.
\end{theorem}
gives a strong bound near the optimal of multi-arm bandit. In order to utilize the advantage of stochastic gradient decent, we sample from the distribution $\pi_t^i$ at every epoch instead of always picking batch with largest weight $max_t{w_t}$ greedily. In spite of simple formulation, our method are proved to be very reasonable.

\subsection{Strategy of Sampling anchor points}
Detection of nodules in CT face large imbalance between positive and negative examples, over 10000: 1 in our task. We discover learning negative samples in an easy to hard manner makes the training procedure more stable. During training, we sample negative samples which confidence is between a threshold $[\xi_i, \eta_i]$ from every batch at task $\mathcal{T}_i$. Both $\eta_i$ and $\xi_i$ gradually descends linearly with steps of training and finally approaching to $[0, \eta_{End}]$ to make the task more and more challenging through time. $\eta_{End}$ is a threshold select for hard samples, which in practice we choose 0.3.

\renewcommand{\algorithmicrequire}{\textbf{Input:}}
\begin{algorithm}
\caption{Curriculum Training Strategy}
\label{alg:1}
\begin{algorithmic}
\REQUIRE $\mathcal{X} = \{(x_1, y_1), ... (x_n, y_n)\}$, $\mathcal{S} = \{(L_1, N_1), ... (L_n, N_n)\}$
\FOR{$i=0$ to $N$}
\STATE {
Calculate weight for every sample in $\mathcal{X}$ by $w_t := L_t + \frac{\alpha}{\sqrt{N_t}}$\\
Rescaling weights to [0, 1]\\
Computing Sample distribution $D_i$ by $\pi_t^i := (1 - \epsilon)\frac{e^{w_t}}{\sum_{j=1}^n{e^{w_j}}} + \frac{\epsilon}{n}$\\
Sample $N_{epoch}$ data from $D_i$ and group them into batch $\mathcal{B} = \{b_1, ..., b_k\}$
\FOR{$b_j$ in $\mathcal{B}$}
\STATE {
Sample negative anchors between threshold $[\xi_i, \eta_i]$.\\
Train the network on all positive anchors and sampled negative anchors\\
Update $(L_t, N_t)$ for samples in this batch
}
\ENDFOR
}
\ENDFOR
\end{algorithmic}
\end{algorithm}

\section{Experiments}
\vspace{-0.2cm}
\subsection{Experiment Settings}
CTs are normalized with setting window level to -600 and window width to 1600. We resize CTs to pixel spacing 0.8 for proposal network. The anchors size for FPN is chosen to be [$4^3, 8^3, 16^3, 32^3$] at different feature map. We use sliding window with window size $128^3$ in proposal stage because the limitation of GPU memory. For image pyramid, we choose pixel spacing 1.0 for large proposals and 0.5 for small ones. $\alpha$ and $\epsilon$ in Alg. \ref{alg:1} is set to 2 and 0.2 respectively, $N_{epoch}$ is set as 0.1x size of the training set.

\vspace{-0.4cm}
\subsection{LUNA16}
\vspace{-0.2cm}
We compared our results with the top three on LUNA16\cite{DBLP:journals/corr/SetioTBBBC0DFGG16} and two published methods\cite{dou2017automated, ding2017accurate}. We further analyze the top 100 false positives which mainly caused by: (1) very ambiguous nodule-like area; (2) the predicted center near but out of the nodule; (3) very obvious false positives. We found that (1) occupies half of our top 100 false positives in the experiment.
\vspace{-0.8cm}
\begin{table}
    \centering
    \caption{LUNA16 Results}
    \begin{tabular}{l|ccccccc|l}
        \toprule
        Teams                   & 0.125 & 0.25  & 0.5   & 1 & 2 & 4 & 8 & Mean\\
        \midrule
        Qi Dou, et al.\cite{dou2017automated}
        & 0.659 & 0.745 & 0.819 & 0.865  & 0.906 & 0.933 & 0.946 & 0.839\\
        Jia Ding, et al.\cite{ding2017accurate}
        & 0.748	& 0.853	& 0.887	& 0.922	 & 0.938 & 0.944 & 0.946 & 0.891\\
        Patech(1st)\cite{DBLP:journals/corr/SetioTBBBC0DFGG16}	        & 0.908	& 0.921	& 0.935	& 0.957	 & 0.97	 & 0.981 & 0.985 & \textbf{0.951}\\
        JianpeiCAD(2nd)\cite{DBLP:journals/corr/SetioTBBBC0DFGG16}	    & 0.884	& \textbf{0.94}	& \textbf{0.96}	& 0.962	 & 0.965 & 0.967 & 0.968 & 0.95\\
        FONOVACAD(3rd)\cite{DBLP:journals/corr/SetioTBBBC0DFGG16}	& \textbf{0.91}	& 0.932 & 0.945	& 0.953	 & 0.96	 & 0.963 & 0.965 & 0.947\\
        \textbf{Our Method}     & 0.811 & 0.9	& 0.948	&\textbf{0.979}  & \textbf{0.984} & \textbf{0.986} & \textbf{0.986} & 0.942\\
        \bottomrule
    \end{tabular}
\end{table}
\vspace{-1cm}
\subsection{NLST and LIDC-IDRI}
The National Lung Screening Trial (NLST) was a randomized controlled clinical trial of screening tests for lung cancer\cite{national2011reduced}. We labeled the dataset with 4 radiologists using labeling procedure descripted in \cite{armato2011lidc}. Total 2836 CT and 4595 nodules are labeled. LUNA16 is originated from LIDC-IDRI\cite{armato2011lidc} but excludes scans with slice thickness greater than 2.5 mm and nodules $<$ 3mm. As we are interested in small nodules, we include 801 CTs from LIDC which consists 4610 nodules, among which 2837 are around or under 3mm. As a whole, our training dataset consists of total 5673 CTs and 9206 nodules. We split both dataset randomly into 80\%, 10\%, 10\% as training, validation and testings sets respectively.

\begin{figure}[ht]
\centering
\includegraphics[width=\columnwidth]{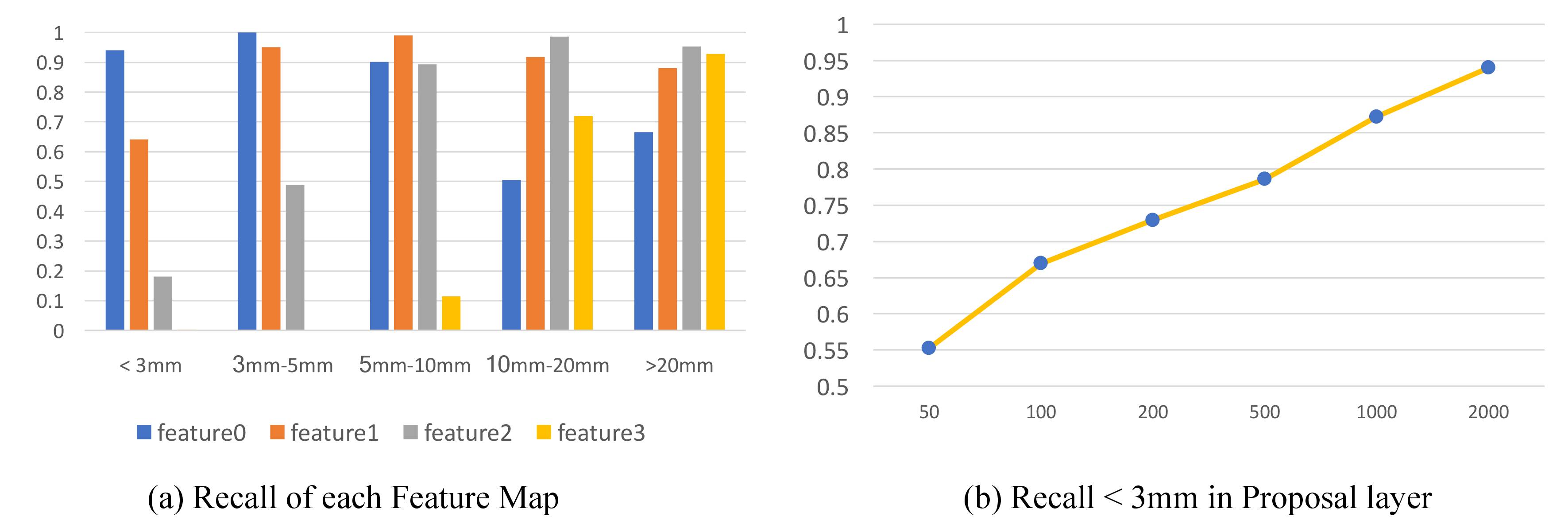}
\caption{Some Results on NLST and LIDC}
\label{fig2}
\enspace
\end{figure}

We give our results of LIDC-IDLR, the testing set consists of 80 CTs and 431 nodules, 266 nodules of which are $<= 3mm$. Our method reached \textbf{100\%} recall for nodules $>3mm$ when picking the top 50 proposals at each feature map. For nodules $<=3mm$, the relation between recall and proposal number is in Fig. \ref{fig2} (b). For final results on LIDC, we remain \textbf{100\%} recall for test sets, and reach \textbf{80.36\%} for nodules $< 3mm$ at 11 FP/s. As it is hard to classify 2000 proposals in one step, the result is achieved by cascade two false positive network. For NLST dataset, we finally reach \textbf{97.53\%} with about 7 false positives per scan.\\
\vspace{-0.8cm}
\begin{table}[]
    \centering
    \caption{Nodule Recall of different size}
    \begin{tabular}{l|cccc}
    \toprule
    Recall &     $<3mm$ & 3mm-5mm & 5mm-10mm & $>10mm$  \\
    \midrule
    NLST   &    1.0   & 1.0     & 96.28  &   96.64 \\
    LIDCC  &    84.58 & 1.0     & 1.0    &    1.0\\
    \bottomrule
    \end{tabular}
    \label{tab:my_label}
\end{table}
\vspace{-1cm}

To evaluate the effectiveness of the 3D FPN, we test each feature map's recall of different size of nodules. Result is shown in Fig. 2. We can clearly observe a trend that small nodules comes in lower layers while the large ones acts oppositely. This trend provides a strong guarantee for the validity of our integration of feature and image pyramid.

\subsection{Time of Training}
We tested the training time on two datasets: (1) LUNA16; (2) combination of LIDC and NLST. We record time of reaching average training accuracy on sampled anchor at 85\%, 90\%, 95\% and 98\% respectively. We tested with on 4 Titan Pascal X on PyTorch with batch size as 4.
\vspace{-0.4cm}
\begin{table}
    \centering
    \caption{Training time of our Proposal Network}
    \begin{tabular}{l|cccc}
        \toprule
        Datasets                       & 85\%  & 90\% & 95\%  & 98\%\\
        \midrule
        LUNA16(without curriculum)     & 2.1h  & 3.8h & 8.4h  & 21.7h\\
        LUNA16(with curriculum)        & 2h    & 2.8h & 5.3h  & 11.5h  \\
        NLST+LIDC(with curriculum)     & 6.3h  & 8h   & 13.4h & 30.2h \\
        \bottomrule
    \end{tabular}
\end{table}

\section{Conclusion}
In this paper, we proposed a novel detector which efficiently integrates Feature and Image Pyramid. Our network is strong in handling large variation of size and achieve state of art performance on LUNA16 dataset (average 94.2\%). Meanwhile, to faster training, we propose a general curriculum training strategy which halves the training time of our proposed network. We show that our proposed system is able to work on larger datasets.
\bibliographystyle{splncs}
\bibliography{ref}

\end{document}